\documentclass[letterpaper]{article} 
\usepackage{aaai24}  
\usepackage{times}  
\usepackage{helvet}  
\usepackage{courier}  
\usepackage[hyphens]{url}  
\usepackage{graphicx} 
\usepackage{natbib}  
\usepackage{caption} 
\usepackage{algorithm}
\usepackage{multirow}
\usepackage{makecell}
\usepackage{pifont}
\usepackage{bbding}
\usepackage{amsmath}
\usepackage{amssymb}
\usepackage{algpseudocode}
\usepackage{booktabs}
\usepackage{amstext}
\usepackage{bm}
\newcommand{\eg}{\textit{e}.\textit{g}.}
\usepackage{enumitem}
\usepackage[table]{xcolor}

\usepackage{url}            
\usepackage{booktabs}       
\usepackage{mathtools,amssymb}
\usepackage{amsfonts}       
\usepackage{nicefrac}       
\usepackage{microtype}      
\usepackage{pgfplots,pgfplotstable}
\pgfplotsset{compat=1.14}
\usepackage{array,colortbl}
\usepackage{xcolor}
\usepackage{algorithm,algorithmicx,algpseudocode}
\usepackage[capitalise]{cleveref}
\usepackage{caption}
\usepackage{graphbox}
\usepackage{placeins}
\usepackage{subcaption}
\usepackage{etoolbox}

\newcommand{\bx}{\mathbf{x}}

\newcommand{\bz}{\mathbf{z}}

\newcommand{\bepsilon}{{\boldsymbol{\epsilon}}}

\newcommand{\grad}{\nabla}

\begin{document}
\title{Generating and Reweighting Dense Contrastive Patterns\\for Unsupervised Anomaly Detection}
\author{
    Songmin Dai\textsuperscript{\rm 1}\thanks{The first two authors contributed equally to this paper.},
    Yifan Wu\textsuperscript{\rm 1}\footnotemark[1],
    Xiaoqiang Li\textsuperscript{\rm 1}\thanks{Corresponding author.},
    Xiangyang Xue\textsuperscript{\rm 2}
}
\affiliations{
    \textsuperscript{\rm 1}School of Computer Engineering and Science, Shanghai University\\
    \textsuperscript{\rm 2}School of Computer Science, Fudan University\\
    laodar@shu.edu.cn, VictorWu@shu.edu.cn, xqli@shu.edu.cn, xyxue@fudan.edu.cn
}

\maketitle

\begin{abstract}
Recent unsupervised anomaly detection methods often rely on feature extractors pretrained with auxiliary datasets or on well-crafted anomaly-simulated samples. However, this might limit their adaptability to an increasing set of anomaly detection tasks due to the priors in the selection of auxiliary datasets or the strategy of anomaly simulation. To tackle this challenge, we first introduce a prior-less anomaly generation paradigm and subsequently develop an innovative unsupervised anomaly detection framework named GRAD, grounded in this paradigm. GRAD comprises three essential components: (1) a diffusion model (PatchDiff) to generate contrastive patterns by preserving the local structures while disregarding the global structures present in normal images, (2) a self-supervised reweighting mechanism to handle the challenge of long-tailed and unlabeled contrastive patterns generated by PatchDiff, and (3) a lightweight patch-level detector to efficiently distinguish the normal patterns and reweighted contrastive patterns. The generation results of PatchDiff effectively expose various types of anomaly patterns, e.g. structural and logical anomaly patterns. In addition, extensive experiments on both MVTec AD and MVTec LOCO datasets also support the aforementioned observation and demonstrate that GRAD achieves competitive anomaly detection accuracy and superior inference speed.
\end{abstract}

\section{Introduction}
\label{sec:introduction}

\begin{figure}[!t]
    \centering
    \includegraphics[width=1\linewidth]{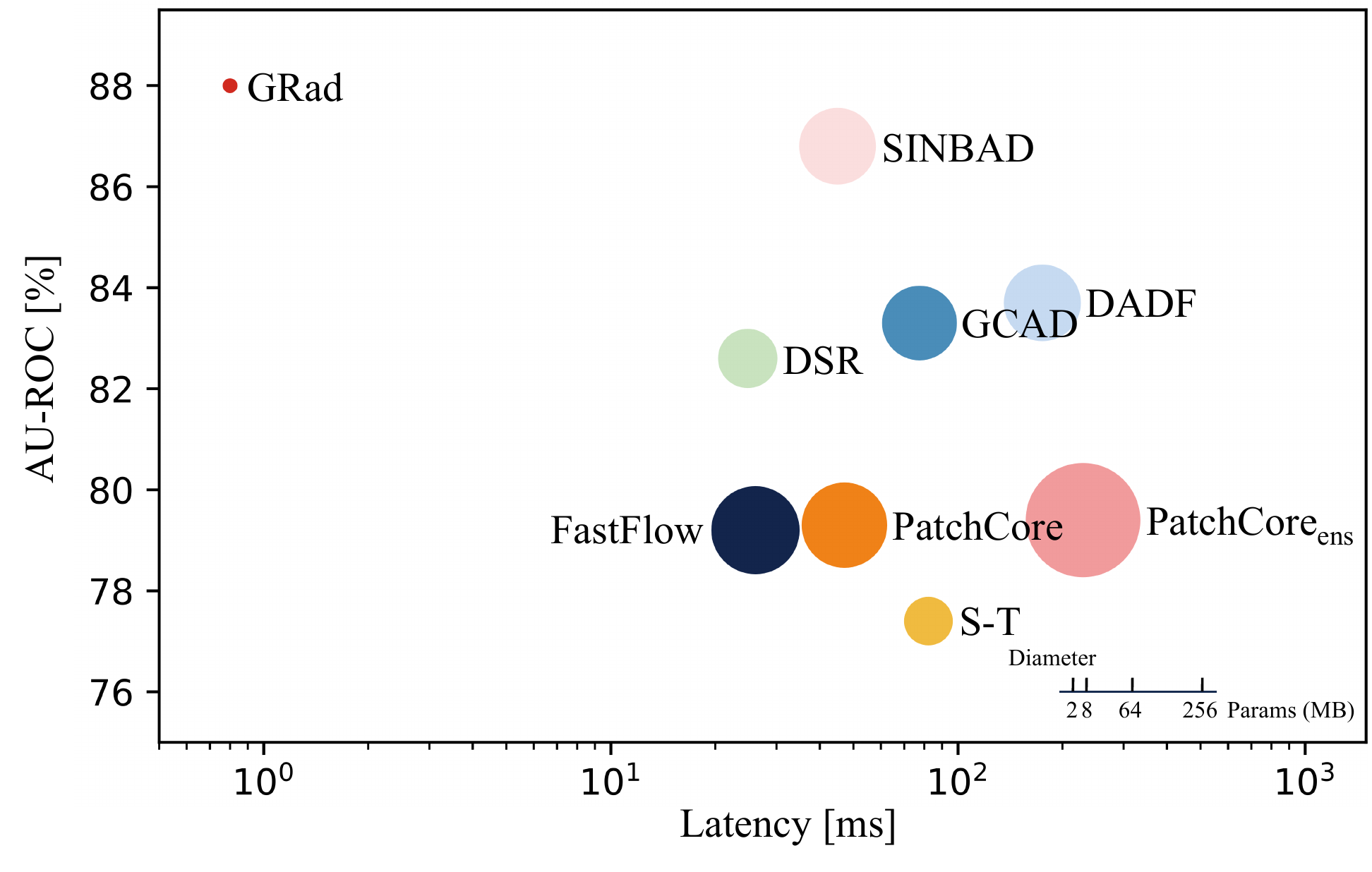}
    \vspace{-0.8cm}
    \caption{Anomaly detection performance vs. latency per image on an NVIDIA Tesla V100 GPU. Each bubble’s area is proportional to the number of parameters in each detector, and each AU-ROC value is an average of the image-level detection AU-ROC values on MVTec LOCO~\cite{MVloco}.}
    \label{fig: speed}
    \vspace{-0.6cm}
\end{figure}

Image anomaly detection plays a crucial role in various fields, including industrial product defect detection, medical image lesion detection, security screening using X-ray images, and video surveillance~\cite{ComplentaryGAN, MVTecAD, app1, app2, GANomaly}. However, securing real-world anomalous data for training is typically challenging and scarce due to the inability to cover a sufficiently diverse range of potential anomaly patterns. 
Consequently, the setting of one-class learning, which employs only normal samples for model training, has proven to be better suited for most industrial anomaly detection tasks~\cite{MVTecAD, MVloco}. 
In recent years, many high-accuracy industrial anomaly detection methods heavily rely on ImageNet~\cite{ImageNet10} pretrained feature extractor. Nevertheless, such reliance may limit their generalization capabilities in scenarios~\cite{MVloco} where ImageNet pretrained features are insufficiently informative, or on other types of image-like data~\cite{mvtec3D, BackTo3dFeatures}. Additionally, some methods have achieved promising results on the MVTec AD~\cite{MVTecAD} without using pretrained feature extractors. These methods utilize manually-selected external out-of-distribution (OOD) datasets~\cite{FCDD} or carefully designed anomaly-simulated data to sample anomaly patterns~\cite{CutPaste, DRAEM, SLSG}. However, previous anomaly acquisition strategies can be considered as ad-hoc solutions that overly rely on priors or visual inspection of test images, such as in MVTec AD, where most anomalies are low-level structure anomalies (\eg, scratches, dents, and contaminations). Such reliance may cause these strategies to fail in detecting other types of anomalies, such as logical anomalies recently proposed in the MVTec LOCO~\cite{MVloco}. These logical anomalies are represented as violations of logical constraints in images, which not only challenges the ananomaly simulation-based methods but also the pretrained representations by auxiliary datasets. Therefore, it becomes necessary to devise image anomaly detection techniques that are independent of both pretrained Imagenet feature extractors and ad-hoc anomaly acquisition strategies.

\begin{figure}[!t]
    \centering
    \includegraphics[width=\linewidth]{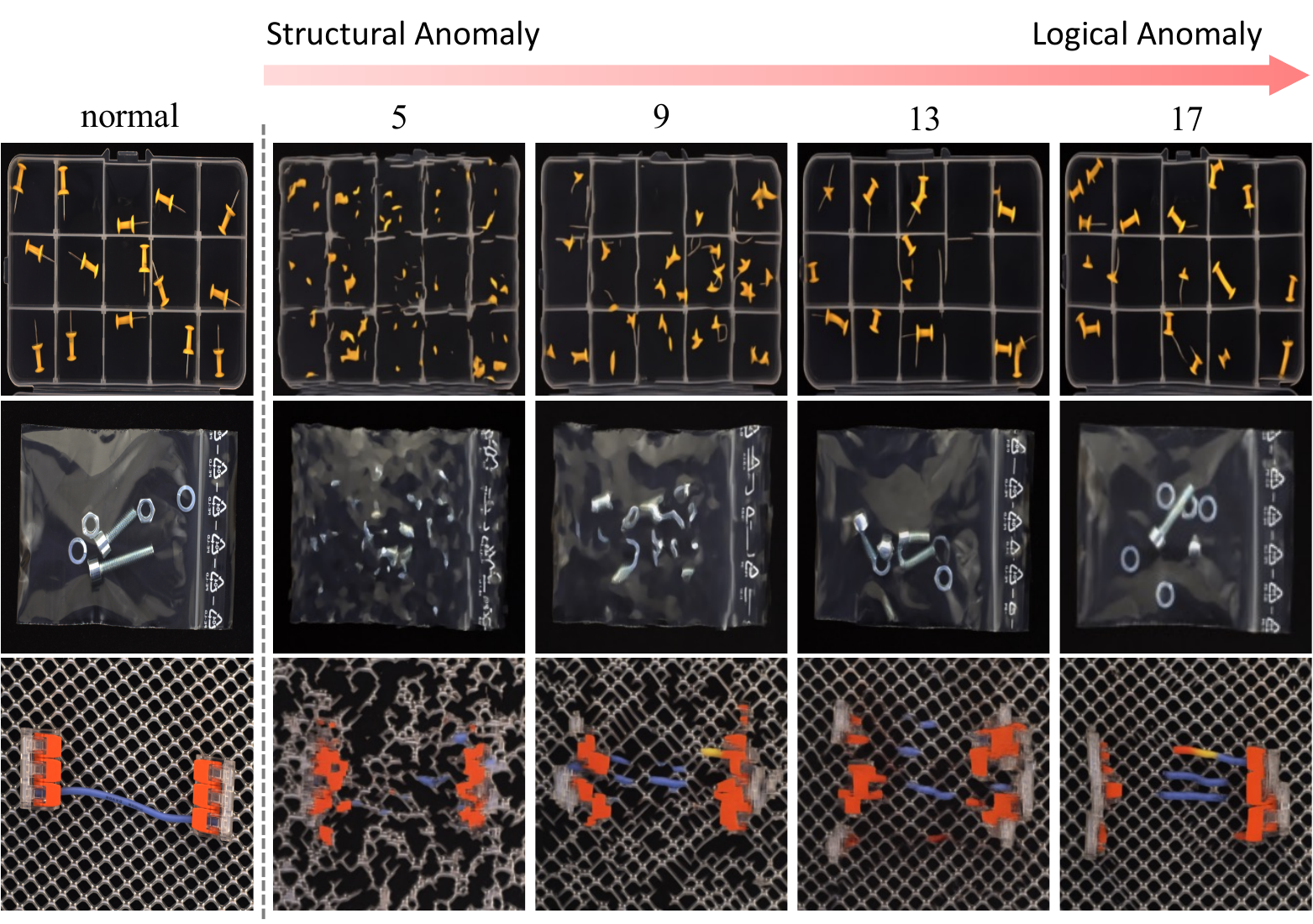}
    \vspace{-0.4cm}
    \caption{Anomaly contrastive images generated by our PatchDiff on MVTec LOCO. The number $n$ above the images indicates that this column is generated based on the corresponding $n \times n$ receptive field size. We show that employing varying sizes of limited receptive fields effectively enables the PatchDiff to expose anomalies at different levels: generators with smaller sizes tend to expose structural anomalies, while generators with larger sizes tend to expose logical anomalies.}
    \label{fig: loco_generation_results}
    \vspace{-0.4cm}
\end{figure}

In this paper, we introduce a novel framework named \textbf{GRAD} (\textbf{G}enerating and \textbf{R}eweighting dense contrastive patterns for unsupervised \textbf{A}nomaly \textbf{D}etection), which achieves SOTA performance in both anomaly detection accuracy and inference runtime, as depicted in Fig.~\ref{fig: speed}. We first put forward a novel anomaly generation paradigm: retaining the structure information within each small patch of the image while disregarding the global structure information of the whole image. Based on this paradigm, we design an anomaly generator called PatchDiff. This generator enforces a constraint on the receptive field size of the diffusion model~\cite{DDPM} and removes the attention layers~\cite{AttentionNotNeed}, thus ensuring that only the local structure within each patch is retained, while the global structure is discarded. As illustrated in Fig.~\ref{fig: loco_generation_results}, with different sizes of the receptive field, PatchDiff can generate diverse dense contrastive patterns that cover a range of anomaly types, \eg, the structural and logical anomalies proposed in MVTec LOCO. Subsequently, we expect to utilize the generated local anomaly patterns to learn a patch-level anomaly detector. However, the contrastive patterns generated by PatchDiff may also be normal and we cannot provide patch-wise ground truth for them. Consequently, the generated contrastive patterns are unlabeled. Furthermore, the local patterns in both normal and generated data could often be long-tailed. Considering the previous two points, we introduce a self-supervised reweighting mechanism to mitigate the negative impacts of fake anomalous patches (patches without effective anomaly patterns) and imbalanced distribution. The mechanism utilizes density information of the features extracted by the detector during the training phase to assign different weights to the contrastive patches. It filters the fake anomalous patches and rebalances the distribution of the contrastive patches. Finally, to obtain high-throughput anomaly detection models better applied in practical industrial scenarios, we design a lightweight Fully Convolutional Network (FCN)-based patch-level detector with a pure encoder architecture. It consists of only 8 convolutional layers but performs on par with larger models in industrial anomaly detection. Furthermore, to deal with tasks that involve mixed-level anomalies, we can also integrate multiple detectors with different receptive fields. We empirically find that a single-level detector is enough to achieve competitive accuracy on MVTec AD dataset, while three detectors can be integrated to handle both structural and logical anomalies in MVTec LOCO.

The main contributions of this paper can be summarized as follows:
\begin{itemize}
    \item We propose a novel paradigm for generating anomaly patterns without scenario-specific priors. Based on this, we develop PatchDiff which can effectively expose a range of local anomaly patterns.
    \item We introduce a self-supervised reweighting mechanism for the generated contrastive data to rebalance them and filter out the fake anomalous patches. This mechanism enables we can efficiently use the unlabeled and long-tailed contrastive patterns for anomaly detection.
    \item We design a lightweight encoder-based patch-level detector trained with only the normal data and generated contrastive data, which relies on no external dataset, heavy pretrained backbone, or memory-consuming decoder architecture. 
\end{itemize}

\section{Related Works}

\begin{figure*}[!t]
\centering
\includegraphics[width=\linewidth]{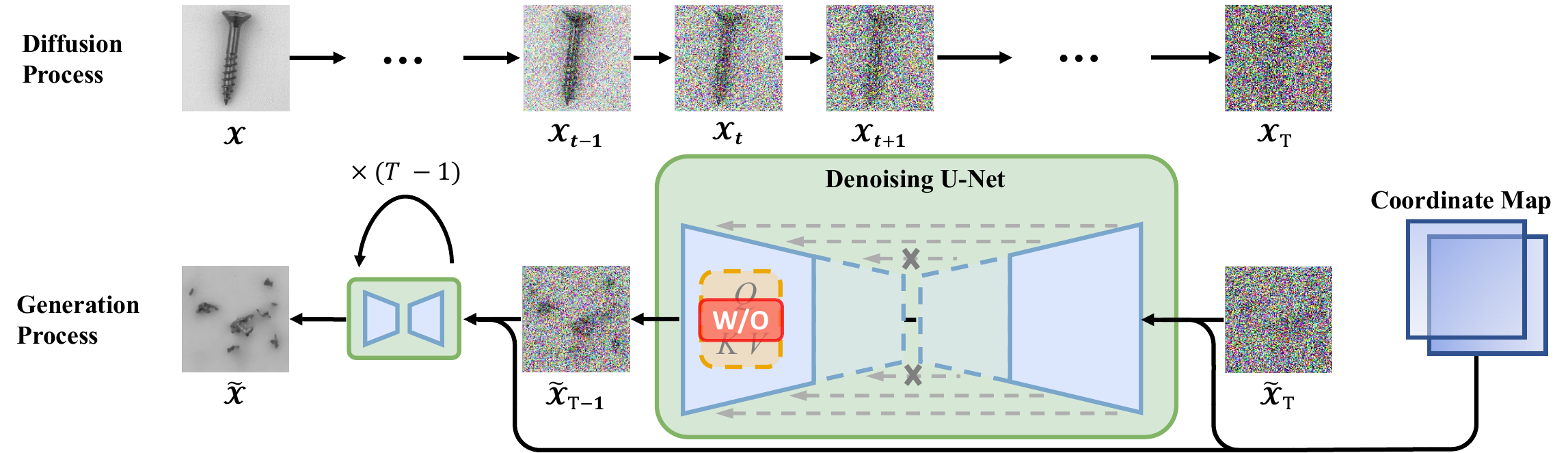}
    \caption{An illustration of the proposed anomaly generator, PatchDiff. Compared with usual Diffusion models, PatchDiff limits the receptive field of the U-Net used to denoise, which preserves only the local structures rather than the global structures. PatchDiff can effectively produce higher-level novel visual structures coming from the recombinations of specific-level local structures. We can use PatchDiffs with various receptive filed sizes to generate multilevel dense contrastive patterns, which are useful for exposing the multilevel anomalies like the structure anomaly and the logical anomaly in MVTec LOCO.}
    \label{fig: PatchDiff}
\end{figure*}

\noindent\textbf{Reconstruction-based.}
A well-trained autoencoder (AE) on normal data is supposed to produce lower reconstruction errors on the normal data than the anomalous data~\cite{AE1, AE2, VAE}. However, in practice, it may also reconstruct anomalies very well or even better~\cite{Pidhorskyi2018GenerativePN}. To alleviate this problem, recent works developed many advanced variants of AE by using generative priors or novel architectures~\cite{OCGAN, MemAE, DAAD, RIAD, InTra, SSPCAB, UniAD}.

\noindent\textbf{Pretrained feature-based.}
State-of-the-art methods for industrial anomaly detection tend to use features of a deep network pretrained on external datasets (\eg, ImageNet). These methods~\cite{PaDiM, DifferNet, Cflow-ad, PatchCore, hyun2023reconpatch, zhang2023prototypical} effectively utilize the general low-level visual features encoded by the pretrained network to do the anomaly detection and achieve appealing performance on MVTec AD~\cite{MVTecAD}. However, they are hard to directly apply in other image-like domains (e.g. the depth map)~\cite{mvtec3D, BackTo3dFeatures} or to cover the higher-level anomaly type, logical anomalies~\cite{MVloco}.

\noindent\textbf{Anomaly simulation-based.}
To overcome the limitations of pre-trained features and ensure that the model produces well-defined and expected results outside the normal distribution, several anomaly simulation methods~\cite{FCDD, CutPaste, DRAEM, SLSG} are proposed. They employ various ad-hoc strategies to simulate specific types of anomaly patterns tailored to different datasets. Most of them heavily rely on human priors and can only expose specific anomaly patterns, making them also challenging to generalize to different scenarios.

\section{Approaches}

Our method can be primarily divided into two stages: (1) Generating diverse contrastive images based on our novel proposed anomaly generation paradigm to cover the anomaly patterns at interest levels. (2) Training lightweight patch-level detectors with our proposed reweighting mechanism to fully utilize the unlabeled and long-tailed generated contrastive patterns. In the following, we will describe the key parts of GRAD in detail.

\subsection{Generating anomaly Contrastive Images}
\label{sec: generating}

In contrast to previous ad-hoc anomaly acquisition strategies~\cite{CutPaste, DRAEM, SLSG}, we introduce a novel and prior-less anomaly generation paradigm: preserving the structure information within each small image patch while disregarding the global structure information of the entire image. To implement this, we propose a diffusion model~\cite{DDPM} based generator called PatchDiff.
As shown in Fig.~\ref{fig: PatchDiff}, the diffusion and denoise process is very similar to DDPM, the differences mainly come from the U-Net architecture in the following aspects:  

(1) To prevent the U-Net from utilizing long-range information for recovering global structures during denoising, we deliberately remove self-attention used in DDPM~\cite{DDPM}. Self-attention is a powerful tool for capturing long-range contextual information, but for our specific task, it is unnecessary~\cite{AttentionNotNeed}, since local consistency is all we need. 

(2) To further ensure that the U-Net focuses on recovering the local patterns within the corresponding patches during denoising, we directly reduce the depth of both the encoder and decoder of the U-Net. In this way, each latent neuron of the bottleneck has a limited receptive field, and thus it denoises using only the local content and retaining only local structures. 

(3) To enable the U-Net to effectively model position-dependent cues, we incorporate a 2-channel coordinate map as additional information alongside the input. This coordinate map is a tensor with dimensions matching that of the input image, where each element represents the coordinate of the corresponding pixel. Noteworthy, the output of our U-Net is still a 3-channel image as same as the original U-Net in DDPM.

Then we modify the training loss of original Diffusion models by introducing a global tiny noise $\bepsilon_{g}$ during the noise injection process. It is motivated by the observation that there is a tendency for overall color deviation in the generated results. Consequently, to avoid the color deviation, the training loss of PatchDiff at each denoising step $t$ becomes 
\begin{equation*}
\begin{aligned}
\mathbb{E}_{\bepsilon_1, \bepsilon_g}
\big\| \bepsilon_1 - \bepsilon_\theta\bigl(\sqrt{\bar{\alpha}_t} \mathbf{x}_0 + \sqrt{1-\bar{\alpha}_t} \bepsilon_1 + \bepsilon_g, t\bigr) \big\|^2,
\end{aligned}
\end{equation*}
where $\epsilon_{1}\sim\mathcal{N}(\mathbf{0}, \mathbf{I})$, $\epsilon_{\theta}$ and $\bar{\alpha}_t$ are the same as in DDPM. 
As depicted in Fig.~\ref{fig: loco_generation_results}, the images generated by PatchDiff effectively avoid the presence of low-level anomalous cues that often occur in simulation strategies, easily noticeable edges when tailoring two images together. 
Instead, PatchDiff focuses more on the slightly higher-level anomaly patterns. By setting multiple receptive filed sizes to the U-Net, PatchDiff can efficiently expose both structural and logical anomalies in MVTec LOCO. This enables PatchDiff to produce more comprehensive local abnormalities without using any prior knowledge of test anomalies. Additionally, the reduction in the depth of architecture and the removal of attention layers contribute to a decrease in the model's complexity and calculation cost, leading to improved training and sampling speed. Furthermore, it is worth noting that the training process of PatchDiff uses only fitting loss like DDPM\cite{DDPM}, which is very stable and easy to implement. In summary, PatchDiff is a prior-less, easy-to-implemented, relatively-fast multilevel local anomaly pattern generation method. 

\begin{figure}[!t]
\centering
    \includegraphics[width=1\linewidth]{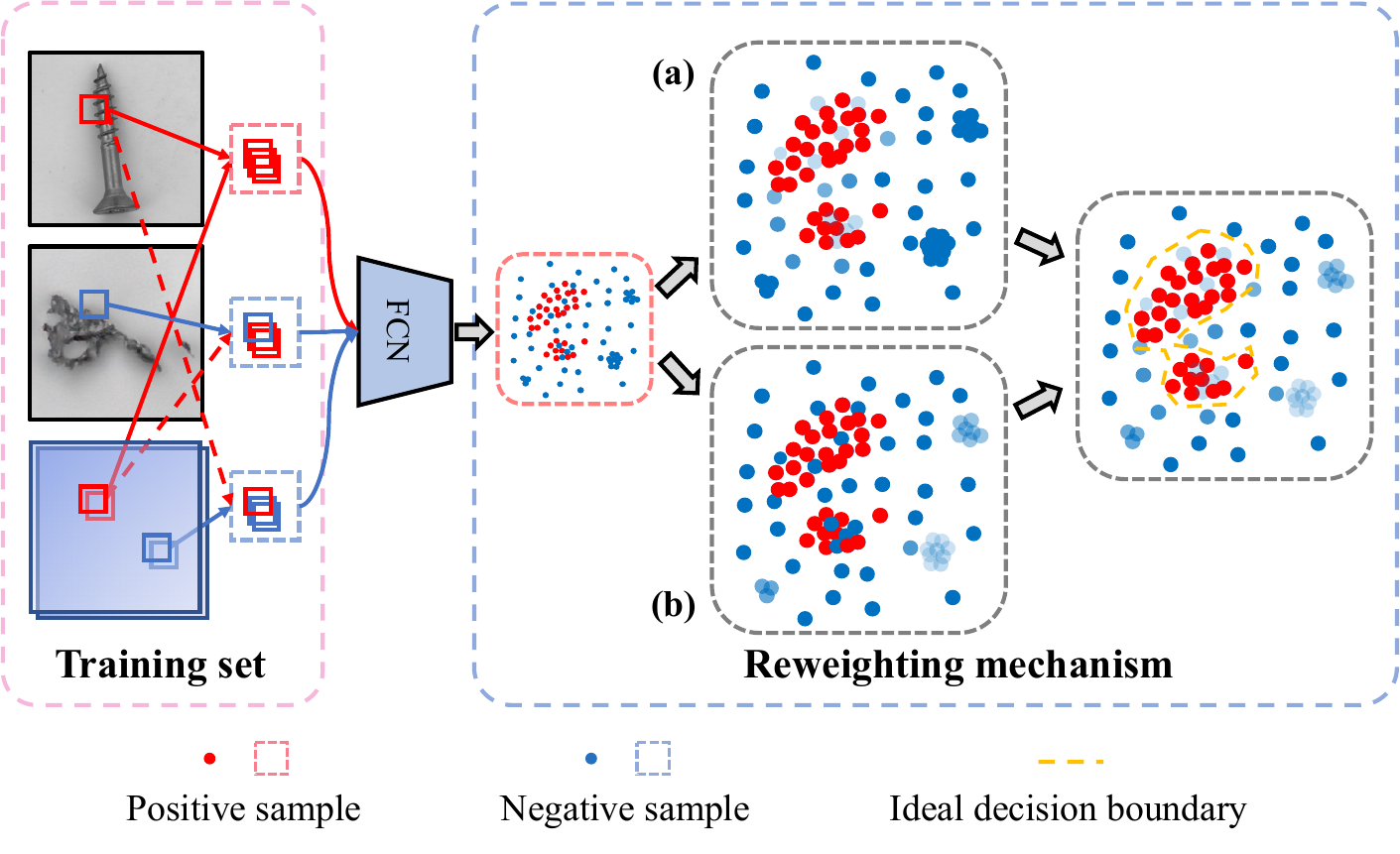}
    \vspace{-0.6cm}
    \caption{Schematic overview of two components during training patch-level detectors. The left portion is the training set which consists of one type of positive patch and two types of negative patches. The right portion is the reweighting mechanism which comprises mechanism (a) to filter the fake anomaly patterns and mechanism (b) to rebalance the long-tailed training data.}
    \label{fig: detector}
    \vspace{-0.4cm}
\end{figure}

\subsection{Training Patch-level Detector}
\label{sec: training}
A naive idea to utilize the contrastive images generated by PatchDiff is directly labeling them as the anomalous class and training image-level detectors. But it does not fully exploit the dense and local anomaly patterns nor provide useful anomaly scores for localization. Instead, we opt to train patch-level anomaly detectors that detect level-specific local anomalies by patch-wisely classifying the normal images and contrastive images. 
Our patch-level detector is implemented with an 8-layer Fully Convolutional Network, FCN~\cite{FCN}, in a pure encoder way. At the training stage, the detector takes input patches of a fixed size, precisely $34 \times 34$ pixels, and produces an output anomaly score corresponding to each individual patch. To address local anomalies of multiple concerned levels (\eg, both structural and logical anomalies in MVTec LOCO), we choose to maintain the detector architecture but resize the original images into lower resolutions, which indirectly achieves the adjustment of the receptive field sizes. This approach enables us to train additional detectors capable of capturing higher-level anomaly patterns without redesigning the detector's architecture and further reduces the computational cost. In the following, we will introduce how to train the patch-level detector.

\begin{figure}[!t]
    \centering
    \includegraphics[width=1\linewidth]{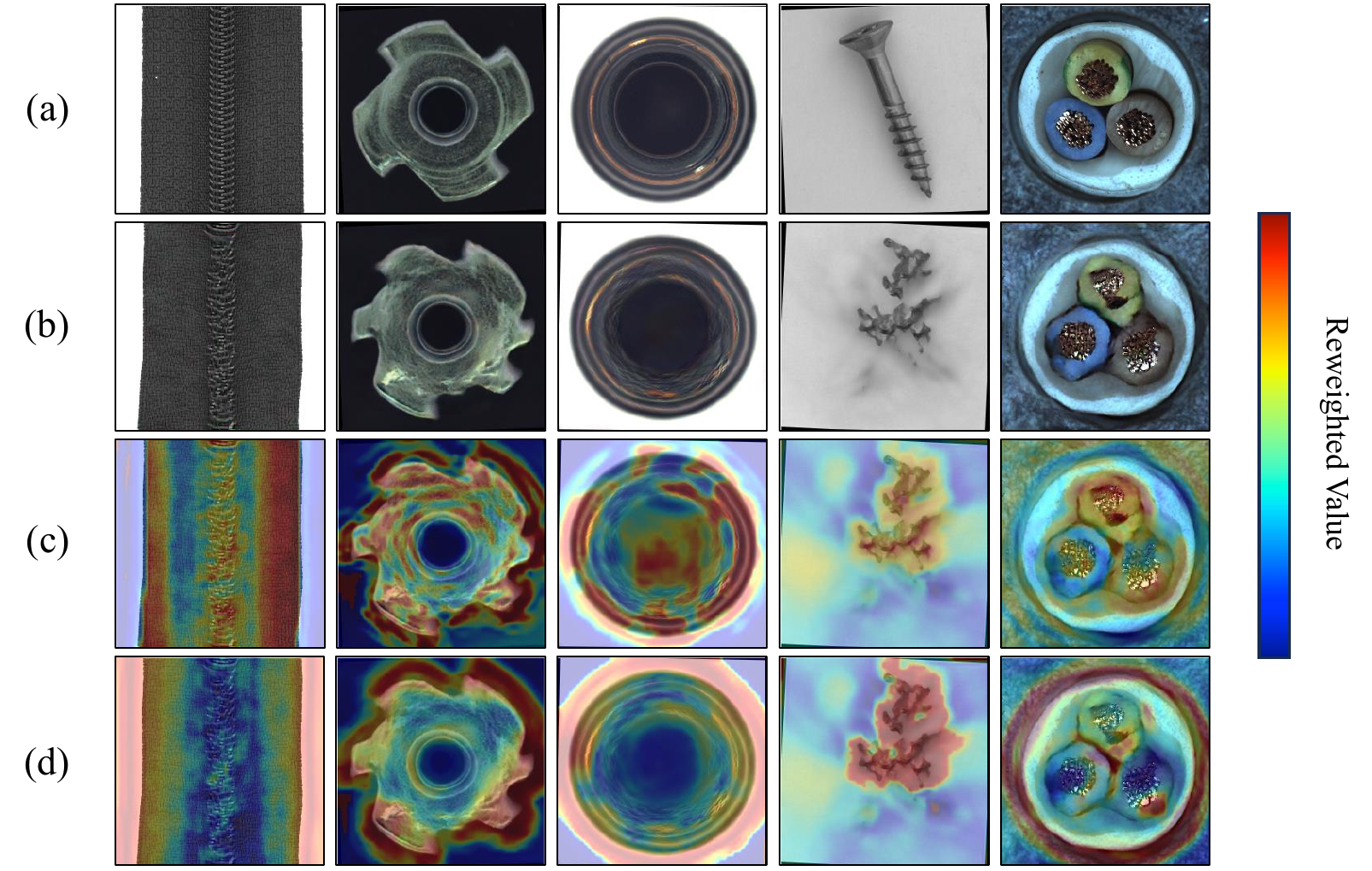}
    \caption{The reweighted map to show the effects of reweighting mechanism. (a) and (b) respectively displays the origin images and the generated contrastive images. (c) and (d) respectively depicts the effects when filtering fake anomaly patterns and rebalancing long-tailed training data. Our reweighting mechanism learns to identify patterns to be disregarded, indicated by the blue regions, and patterns to be emphasized, represented by the red regions, through a self-supervised approach} 
    \label{fig:reweightmap}
\end{figure}

\subsubsection{Preparing the Training set} 

Similar to the input during the generation phase, we use a 2-channel coordinate map $F$ as an additional input. As illustrated in the left portion of Fig.~\ref{fig: detector}, we prepare three types of 5-channel patches as training inputs, including one type of positive patch and two types of negative patches. Let $I$ denote an image data, and $\mathcal{I}^{+}$ and $\mathcal{I}^{-}$ represent sets of normal samples and generated samples from PatchDiff, respectively. Subsequently, the positive patches set $\mathcal{C}^{+}$ and negative patches set $\mathcal{C}^{-}$ are defined as
\begin{equation*}
    \begin{aligned}
    \mathcal{C}^{+}=&\left\{c \mid c=\operatorname{RandCrop}(I\oplus F), I \in \mathcal{I}^{+}\right\},\\
    \mathcal{C}^{-}=&\left\{c \mid c=\operatorname{RandCrop}(I)\oplus \operatorname{RandCrop}(F), I \in \mathcal{I}^{+}\right\} \\
    &\cup\left\{c \mid c=\operatorname{RandCrop}(I\oplus F), I \in \mathcal{I}^{-}\right\},
    \end{aligned}
\end{equation*}
where $\oplus$ denotes concatenation along the channel axis. The negative patches are constructed in two ways: (1) the patches from generated samples along with their corresponding coordinate maps, and (2) the patches from normal samples with incorrect coordinate maps. Specifically, the patches from the latter way are believed to provide examples that break the dependence between patch content and position. This explicitly enhances the detector's utilization of the auxiliary information from the coordinate maps and improves its ability to capture position-aware cues.

\subsubsection{Reweighting the Contrastive Patches}
There are two potential challenges during training the patch-level detector $D$: (1) The images generated by PatchDiff are pixel-unlabeled, leading to the presence of fake anomaly patterns (e.g. the background region in the generated images) among the negative patches, which will mislead the detector. (2) Some important anomaly patterns may appear more rarely and lie in the low-density regions of the data manifold, causing the detector to overlook such patterns during the training process. To mitigate these challenges, we propose a feature density-based reweighting mechanism that incorporates two reweighting strategies, as shown in the right part of Fig.~\ref{fig: detector}. This mechanism relies on the feature distributions extracted from the last latent layer of our patch-level detector on both positive and negative samples. Let us denote $\mathcal{M}^+$ and $\mathcal{M}^-$ as the feature sets of positive and negative samples, respectively. Then the two reweighting strategies can be performed as follows:

(1) Filtering the fake anomaly patterns. As depicted in Fig.~\ref{fig: detector}(a), we introduce a reweighting factor $w^\text{noisy-}_i$ for each given negative patch $\bm{c}^-_i$, to assign smaller weights to the patches whose features are too close to or even within normal features set $\mathcal{M}^+$. The reweighting factor can be formulated as
\begin{equation}
    w^\text{noisy-}_i = \frac{1}{\sum_{\bm{z}\in\mathcal{M}^+} \exp(\beta_\text{density} \text{sim}(\bm{z},\bm{z}^-_i))} ,
\end{equation}
where $\bm{z}^-_i$ is the feature vector of the negative patch $\bm{c}^-_i$, $\mathrm{sim}(\bm{z}, \bm{z}'):= \bm{z} \cdot \bm{z}'/\lVert{\bm{z}}\rVert\lVert{\bm{z}'}\rVert$ is the density kernel based on the cosine similarity and $\beta_\text{density}>0$ is a hyper-parameter for controlling kernel bandwidth. 

(2) Rebalancing the long-tailed training patches. As depicted in Fig.~\ref{fig: detector}(b), we introduce a reweighting factor $w^\text{tail-}_i$ for each given negative patch $\bm{c}^-_i$ to downweight the patches whose features are in the high-density regions. Empirically, we find introducing a reweighting factor $w^\text{tail+}_j$ for each positive patch $\bm{c}^+_j$ is also helpful. Therefore we have the following two additional reweighting factors for the training patches
\begin{equation}
\begin{aligned}
    w^\text{tail-}_i = \frac{1}{\sum_{\bm{z}\in\mathcal{M}^-} \exp(\beta_\text{density} \text{sim}(\bm{z},\bm{z}^-_i))},\\
    w^\text{tail+}_j = \frac{1}{\sum_{\bm{z}\in\mathcal{M}^+} \exp(\beta_\text{density} \text{sim}(\bm{z},\bm{z}^+_j))}.
\end{aligned}
\end{equation}

The effects of our reweighting mechanism are shown in Fig~\ref{fig:reweightmap}. By incorporating these two kinds of reweighting factors, our reweighted binary classification loss $\mathcal{L}_{\mathrm{RBCE}}$ can be formulated as
\begin{equation}
\label{eq: loss_function_rbce}
\begin{aligned}
    \mathcal{L}_{\mathrm{RBCE}}&=-\frac{1}{\lambda^+}\sum_{j=1}^{|\mathcal{C}^+|} w^\text{tail+}_j\log (1-f(\bm{c}^+_j))\\
    &\quad -\frac{1}{\lambda^-}\sum_{i=1}^{|\mathcal{C}^-|} w^\text{tail-}_i w^\text{noisy-}_i \log(f(\bm{c}^-_i)),
\end{aligned}
\end{equation}
where $\lambda^+$ and $\lambda^-$ are the normalization constants to keep the total weights of each class equal to 1:
\begin{equation}
\label{eq: regularization parameter}
    \lambda^+ = \sum_{j=1}^{|\mathcal{C}^+|} w^\text{tail+}_j,\quad
    \lambda^- = \sum_{i=1}^{|\mathcal{C}^-|} w^\text{tail-}_i w^\text{noisy-}_i.
\end{equation}
In practice, the $\mathcal{M}^+$ and $\mathcal{M}^-$ are both implemented with a memory bank that store the features of previous training steps in a queue.

\subsubsection{Regularization on Features and Gradients}
We further utilize a classical unsupervised representation learning method named denoising autoencoder ~\cite{DAE} to regularize the learned feature by detector $D$. To achieve that, we introduce a simple MLP-based network $R$ that recovers the original input patches from the feature vectors extracted from the last latent layer of $D$. Let $f_{Z}$ denote the function extracting features from input patches, $f_{R}$ denote the function recovering input patches from features, and $\mathcal{C}$ denote the collection of all training patches $\mathcal{C}^+\!\cup\!\mathcal{C}^-$. The feature regularization loss can be formulated as
\begin{equation}
    \begin{aligned}
	    \mathcal{L}_{\mathrm{feat}}= \frac{1}{|\mathcal{C}|}\sum_{\bm{c}\in \mathcal{C}} \left\|f_{R}(f_{Z}(\bm{c}+\bm{\epsilon_c})+\bm{\epsilon_z}) - \bm{c}\right\|^2,
     \end{aligned}
\end{equation}
where $\bm{\epsilon_c}$ and $\bm{\epsilon_z}$ are respectively the noise perturbations added to the feature layer and the input layer. The auxiliary denoising task regularizes the last hidden layer of the detector to extract informative and robust representations. We highlight the auxiliary network $R$ will be dropped in the inference stage so that will not increase the inference runtime.

Additionally, we propose a gradient regularization loss to smooth the learned decision function $f$, which further discourages the detector from learning imperceptible distinctions between normal patterns and fake anomaly patterns. The gradient regularization loss can be formulated as
\begin{equation}
    \mathcal{L}_{\mathrm{grad}}=\frac{1}{|\mathcal{C}^+|}\sum_{\bm{c}\in \mathcal{C}^+} \left\|\nabla_{\bm{c}} f(\bm{c})\right\|^{2}.
    \label{eq_gp}
\end{equation}
It penalizes the gradient norms of the decision scores with respect to the input data, which is often used to improve the Lipschitz smoothness and robustness, and thus the generalization performance of decision functions~\cite{GradPU, arjovsky2017towards,inputGrad}. 
\subsubsection{The Overall Training Loss}

We calculate the overall training loss for the patch-level anomaly detector by aggregating the aforementioned three types of losses as
\begin{equation}
    \mathcal{L}=\mathcal{L}_{\mathrm{RBCE}} + \alpha_1 \mathcal{L}_{\mathrm{feat}} +\alpha_2 \mathcal{L}_{\mathrm{grad}},
\label{eq:final_loss}
\end{equation}
where $\alpha_1$ and $\alpha_2$ are hyper-parameters to adjust the impact of $\mathcal{L}_{\mathrm{feat}}$ and $\mathcal{L}_{\mathrm{grad}}$. 

\begin{table*}[!ht]
\centering
    \label{tab:mvtec_main}
    \resizebox{0.95\textwidth}{!}{
    \begin{tabular}{c|c|c|c|c|c|c|c|c}
    \toprule
    Category & \makecell[c]{SPADE\\\tiny{\citealp{SPADE}}} & \makecell[c]{PaDiM\\\tiny{\citealp{PaDiM}}} & \makecell[c]{S-T\\\tiny{\citealp{S-T}}} & \makecell[c]{PatchCore\\\tiny{\citealp{PatchCore}}} &\makecell[c]{GCAD\\\tiny{\citealp{MVloco}}} & \makecell[c]{DADF\\\tiny{\citealp{DADF}}} & \makecell[c]{SINBAD\\\tiny{\citealp{SINBAD}}} & \makecell[c]{GRAD\\\tiny{Ours}} \\ 
    \midrule
    breakfast box 
    & 78.2 & 65.7 & 68.6 & 81.3 & 83.9 & 75.3 & \textbf{92.0} & 81.2 \\
    juice bottle 
    & 88.3 & 88.9 & 91.0 & 95.6 & \textbf{99.4} & 98.6 & 94.9 & 97.6 \\
    pushpins 
    & 59.3 & 61.2 & 74.9 & 72.3 & 86.2 & 81.0 & 78.8 & \textbf{99.7} \\
    screw bag 
    & 53.2 & 60.9 & 71.2 & 64.9 & 63.2 & 77.3 & \textbf{85.4} & 76.6 \\
    splicing connectors 
    & 65.4 & 67.8 & 81.1 & 82.4 & 83.9 & 86.4 & \textbf{92.0} & 85.4 \\
    \midrule
    average 
    & 68.8 & 68.9 & 77.3 & 79.3 & 83.3 & 83.7 & 86.8 & \textbf{87.5} \\
    \bottomrule
\end{tabular}}
\caption{Image-level AU-ROC performance for anomaly detection of different methods on MVTec LOCO~\cite{MVloco}. The best results are in bold.}
\label{tab: auc_mvtec_loco}
\end{table*}

\begin{figure*}[!htbp]
\setlength{\belowcaptionskip}{0.0cm}
\setlength{\abovecaptionskip}{0.1cm}
    \centering
\includegraphics[width=0.9\linewidth]{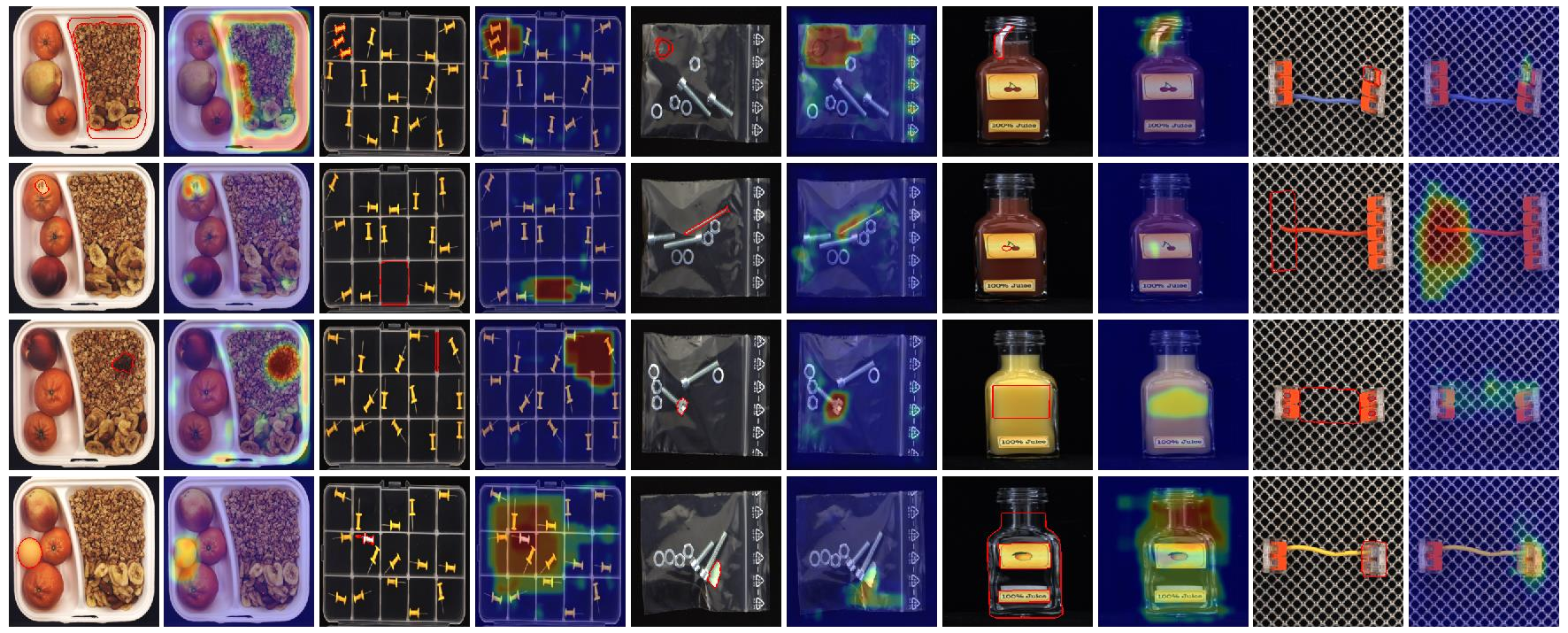}
    \caption{Defect localization results of GRAD on MVTec LOCO~\cite{MVloco}. } 
    \label{fig: main_results}
\end{figure*}

\section{Experiments}

In this section, we first briefly introduce the experimental details (See Appendix for more details). Then we report the anomaly detection accuracies and the ablation study on each component. 

\subsection{Dataset}

To validate the effectiveness and generalizability of our approach, we perform experiments on both MVTec AD~\cite{MVTecAD} and MVTec LOCO~\cite{MVloco}. There are 15 sub-datasets in MVTec AD and 5 sub-datasets in MVTec LOCO and each sub-dataset presents a diverse set of anomalies. 
Particularly, the training sets among them contain only normal images, while the test sets contain both normal and various types of industrial defects. Pixel-level annotations are provided in the test set.

\subsection{Training Settings}

We simply define level-$n$ PatchDiff as the PatchDiff with a receptive field of $n \times n$ pixels, and the images generated by it belong to level-$n$. Similarly, we define level-$n$ detector as the patch-level detector with an indirect receptive field of $n \times n$ pixels.

\noindent\textbf{PatchDiff}. For each sub-dataset in MVTec AD, we train 3 levels of PatchDiffs (level-5, 9, 13). For each sub-dataset in MVTec LOCO, we need to train 3 different levels of detectors, and consequently, we train 4 levels of PatchDiffs (level-5, 9, 13, 17). In particular, 2 of them use level-5, 9, and 13 PatchDiffs and another one uses level-9, 13, and 17 PatchDiffs. For all PatchDiffs, we generally train them for a total of 10,000 training steps. For each sub-dataset, we sample 1,000 images for each level-$n$.

\noindent\textbf{Patch-level Detector}. Each sub-dataset in MVTec AD and MVTec LOCO contains limited training images. To train competitive detectors from scratch for each small sub-dataset, we adopt general data augmentations on both normal and generated images like previous works\cite{MVTecAD, MVloco}. For level-$34, 68,$ and $136$ detectors, the images are respectively resized into $256\times256$, $128\times128$, and $64\times64$. We train the detector on batches of size $128\times(k+2)$ for 2,000 epochs and report the accuracy of the final epoch. Each batch contains 128 randomly cropped positive patches from 4 normal images and $128\times(k+1)$ negative patches from 4 normal images and $4k$ contrastive images, where $k$ equals the number of levels of used generated contrastive images. Specifically, we use $k=3$ for all experiments as mentioned before.

\subsection{Evaluation Settings}
The image-level anomaly score directly takes the max value of a score map from the patch-based anomaly detector, and the pixel-level detection result is obtained by up-sampling the score map and then applying a Gaussian blur with a kernel size of 16. Consistent with existing methods~\cite{MVTecAD, MVloco}, we use AU-ROC as the evaluation metric for the evaluation of image-level anomaly detection and pixel-level anomaly localization.

\begin{table}[htbp]
\centering
\setlength{\belowcaptionskip}{0.0cm}
\setlength{\abovecaptionskip}{0.1cm}
\resizebox{0.9\linewidth}{!}{
\begin{tabular}{lcc}
\toprule
Method & \makecell[c]{Pixel-level\\AU-ROC} & \makecell[c]{Image-level\\AU-ROC} \\ 
\midrule
IGD\tiny{~\citep{IGD}} & 93.1 & 93.4 \\
PSVDD\tiny{~\citep{PSVDD}} & 92.5 & 93.2 \\
FCDD\tiny{~\citep{FCDD}} & 92.1 & 95.7 \\
CutPaste\tiny{~\citep{CutPaste}} & 95.2 & 96.0 \\
NSA\tiny{~\citep{NSA}} & 96.3 & 97.2 \\
DRAEM\tiny{~\citep{DRAEM}} & \textbf{97.3} & 98.0 \\
DSR\tiny{~\citep{DSR}} & - & 98.2 \\
\rowcolor{gray!20} GRAD \tiny{(Ours)} & 96.8 & \textbf{98.7} \\ 
\bottomrule
\end{tabular}}
\caption{Anomaly detection performance on MVTec AD dataset~\cite{MVTecAD}. The best results are in bold.}
\label{tab: auc_mvtec}
\end{table}

\subsection{Main Results}

\noindent\textbf{The anomaly detection results.} We compare GRAD with different methods on MVTec AD and MVTec LOCO, as shown in Table~\ref{tab: auc_mvtec_loco} and Table~\ref{tab: auc_mvtec}. For both datasets, GAD has the best average image-level AU-ROC score, demonstrating the effectiveness of GRAD in anomaly detection. In table~\ref{tab: auc_mvtec_loco}, it is important to note that the fairness of the comparison might be compromised to some extent, as all the compared methods utilize ImageNet pretrained feature extractors. However, GRAD still achieves superior performance by 0.7\% even without such advantages, which shows that ImageNet pretrained features inadequately address the intricacies of logical anomaly detection within MVTec LOCO, and further demonstrates that our contrastive images generated by PatchDiff do expose both structural and logical anomalies effectively. In particular, GRAD achieves excellent results (+13.5\%) on the sub-dataset of pushpins, which exactly fits our observation that the generated images for pushpins perfectly expose several abnormal logical situations in the testing set, \eg, the additional pushpin in the top left compartment and no pushpins in the top right compartment, as shown in level-17 generated pushpin image of Fig.~\ref{fig: loco_generation_results}. In addition, we show the defect localization results in Fig.~\ref{fig: main_results}. In table~\ref{tab: auc_mvtec}, all the methods we compared do not rely on pretrained features and external data. Although GRAD does not achieve the best result for anomaly localization (pixel-level AUROC), it is still competitive among them.

\noindent\textbf{Inference runtimes.} We compare and report the inference latency and FPS in Table~\ref{tab:GRad_speed}. Obviously, GRAD achieves a remarkable throughput performance due to its extremely lightweight architecture, and thereby, GRAD's inference speed is more than 16 times faster than GCAD's.
    
\begin{table}[!htbp]
\centering
\setlength{\belowcaptionskip}{0.0cm}
\setlength{\abovecaptionskip}{0.1cm}
\resizebox{0.9\linewidth}{!}{
\begin{tabular}{lrr}
    \toprule
    Method & latency (ms$\downarrow$) & FPS$\uparrow$ \\
    \midrule
    S-T\tiny{~\cite{S-T}} & 82.2 & 12.2 \\
    FastFlow\tiny{~\cite{FastFlow}} & 26.1 & 38.3 \\
    DSR\tiny{~\cite{DSR}} & 24.8 &40.3 \\
    GCAD\tiny{~\cite{MVloco}} & 12.9 & 77.5 \\
    PatchCore\tiny{~\cite{PatchCore}} &47.1 &21.2 \\
    \rowcolor{gray!20} GRAD \tiny{(Ours)} & \textbf{0.799} & \textbf{1251.6}\\
    \bottomrule
\end{tabular}}
\caption{Inference speed on NVIDIA Tesla V100. The data of our method is obtained on MVTec LOCO dataset with three patch-level detectors (patch size: 34, input size: 256, 128, and 64).}
\label{tab:GRad_speed}
\end{table}

\begin{table}[!htbp]
\centering
\setlength{\belowcaptionskip}{0.0cm}
\setlength{\abovecaptionskip}{0.1cm}
\resizebox{\linewidth}{!}{
\begin{tabular}{lccc}
\toprule
& & \footnotesize{AUROC} & \\ 
& Level-34 & Level-68 & Level-136 \\
\midrule
baseline\dag & 78.2 & 77.8 & 64.3\\
+ Regularization & 81.6 & 80.9 & 65.2\\
+ Noisy Reweighting & 82.5 & 82.5 & 72.1 \\
+ Long-tail Reweighting & 85.2 & 85.4 & 75.1\\
\bottomrule
\end{tabular}}
\caption{Ablation study on components. Detection AUROC results on MVTec LOCO dataset of three patch-level detectors are presented. \dag The baseline setting uses no regularization techniques and reweighting strategies.}
\label{tab: ablation_component}
\end{table}


\subsection{Ablation study}

We first perform an extensive ablation study to validate the effectiveness of two reweighting factors and the regularization technique on MVTec LOCO. The results are shown in Table~\ref{tab: ablation_component}. More details and comprehensive ablation results can be found in Appendix. We utilize the baseline as the beginning and then add regularization, noisy reweighting and long-tail reweighting one by one. 

\textbf{Effects of regularization techniques.} One of the novel contributions presented in this paper is the regularization on features and gradients, which helps our encoder-based detector extract an informative and robust representation and build a smooth decision boundary for the data manifold. As demonstrated in Table~\ref{tab: ablation_component}, the integration of these techniques translates into improvements of +3.4/+3.1/+0.9 on the MVTec LOCO dataset.

\textbf{The effect of reweighting mechanism.} Our reweighting mechanism comprises two essential components: (1) noisy reweighting, which aims to filter fake anomaly patches, and (2) long-tail reweighting, designed to rectify the imbalanced distribution of input data. When integrating the noisy reweighting, our detectors display enhancements of +0.9/+1.6/+6.9 on the MVTec LOCO dataset, as presented in Table~\ref{tab: ablation_component}. Furthermore, with the incorporation of long-tail reweighting, our detectors achieve improvements of +2.7/+2.9/+3.0, as shown in the same table. These outcomes underscore the disruptive influence of fake anomaly patches and the presence of long-tail distributions on detector performance. It is evident that our reweighting mechanism adeptly mitigates these challenges from both fronts, offering substantial advantages to our detectors.

Particularly, in Table~\ref{tab: ablation_component}, Level-136 detectors exhibit relatively poorer performance in anomaly detection. This result can be attributed to their input size, which is merely $64 \times 64$, resulting in insufficient resolution to offer informative structural anomaly details. However, this is in line with our intentions, as the purpose of Level-136 detectors is not to emphasize minute details, but rather to capture the logical relationships among components within the receptive fields of size $136 \times 136$.
    


\section{Conclusion}
In this paper, we propose a novel unsupervised anomaly detection framework, GRAD, by generating and reweighting dense contrastive patterns. The proposed generation method PatchDiff is able to generate multilevel contrastive patterns which exposes a range of local anomaly patterns. The proposed reweighting strategies fully utilize the unlabeled and long-tailed contrastive patterns and help the patch-level anomaly detector better learn the exposed local anomaly patterns. GRAD requires no scenario-specific prior, external datasets, or heavy pretrained feature extractor. It achieves competitive anomaly detection and localization accuracy with a superior inference speed. 

Further refinements to GRAD can be explored: 1) Advancing the algorithms for handling long-tailed and noisy labeled data, thereby utilizing the potential of the generated dense contrastive patterns more effectively. 2) Investigating the feasibility of integrating multiple patch-level detectors into a single lightweight network. 3) Exploring better network architecture and training settings for PatchDiff.

\section{Acknowledgements}

This work is supported in part by Shanghai science and technology committee under grant No. 21511100600.  We appreciate the High Performance Computing Center of Shanghai University, and Shanghai Engineering Research Center of Intelligent Computing System for providing the computing resources and technical support. Additionally, we extend our heartfelt appreciation to Professor Jincheng Jin and Weizhong Zhang from Fudan University, as well as Associate Professor Bin Li, for their invaluable assistance and insights during the writing process of this paper. Their expert guidance and stead- fast support were instrumental in the successful completion of this research.

\bibliography{aaai24}

\onecolumn{
\onecolumn
\section{Appendix}

\subsection{Settings of PatchDiff}
The denoising step $T$ of our PatchDiff is $1000$, and the values of images and positional tensors are normalized into a range of $[-1, 1]$. We use AdamW optimizers with a initial learning rate of 1e-3 and a one-cycle learning rate scheduler. The weight decay strength is set to $0.0001$. The global noise $\epsilon_g$ is sampled from a global Gaussian distribution $mathcal{N}_g$ which has a standard deviation of $\sigma_g=0.02$ and every pixel has the identical noise vector. 

To better demonstrate the training and sampling differences between the PatchDiff and DDPM~\cite{DDPM}, we further present the algorithm details 
of training and sampling in Algorithm 1 and Algorithm 2. In particular, $\bepsilon_\theta$ is a function approximator intended to predict $\epsilon_1$ from $\bx_t$, $\epsilon_g$ is the global noise we introduced, and other variables are the same as in DDPM. Actually the sampling process is completely identical to the DDPM and without affected by the introducing of $\epsilon_g$ in the training stage.
\begin{figure}[!htpb]
\begin{minipage}[t]{0.495\textwidth}
\begin{algorithm}[H]
  \caption{Training} \label{alg:training}
  \small
  \begin{algorithmic}[1]
    \Repeat
      \State $\bx_0 \sim q(\bx_0)$
      \State $t \sim \mathrm{Uniform}(\{1, \dotsc, T\})$
      \State $\bepsilon_{1}\sim\mathcal{N}(\mathbf{0}, \mathbf{I}), \bepsilon_{g}\sim\mathcal{N}_g(\mathbf{0}, \sigma_g\mathbf{I})$
      \State Take gradient descent step on
      \State $\grad_\theta \left\| \bepsilon_1 - \bepsilon_\theta\bigl(\sqrt{\bar{\alpha}_t} \mathbf{x}_0 + \sqrt{1-\bar{\alpha}_t} \bepsilon_1 + \bepsilon_g, t\bigr) \right\|^2$
    \Until{converged}
  \end{algorithmic}
\end{algorithm}
\end{minipage}
\hfill
\begin{minipage}[t]{0.495\textwidth}
\begin{algorithm}[H]
  \caption{Sampling} \label{alg:sampling}
  \small
  \begin{algorithmic}[1]
    \vspace{.04in}
    \State $\bx_T \sim \mathcal{N}(\mathbf{0}, \mathbf{I})$
    \For{$t=T, \dotsc, 1$}
      \State $\bz \sim \mathcal{N}(\mathbf{0}, \mathbf{I})$ if $t > 1$, else $\bz = \mathbf{0}$
      \State $\bx_{t-1} = \frac{1}{\sqrt{\alpha_t}}\left( \bx_t - \frac{1-\alpha_t}{\sqrt{1-\bar\alpha_t}} \bepsilon_\theta(\bx_t, t) \right) + \sigma_t \bz$
    \EndFor
    \State \textbf{return} $\bx_0$
    \vspace{.04in}
  \end{algorithmic}
\end{algorithm}
\end{minipage}
\end{figure}

\subsection{Settings of Patch-based Detectors}
\subsubsection{Data augmentation}
We first present the data augmentation details applied during the training of detectors in MVTec AD and MVTec LOCO, as respectively illustrated in Table~\ref{tab:augmentation_mvtec} and Table~\ref{tab:augmentation_loco}, where $p$ denotes the probability of the images being with color jitter. It is worthy note that a larger color jitter range is applied for each generated set, which is expected to be helpful for learning color-level anomalies without training additional generators and detectors. (In principle, GRAD should expose the color-level structures by directly reduce the receptive field size of PatchDiff to 0 and generate pure noise images, then learn the color-level anomalies by level-1 detector)

\begin{table}[!htbp]
\centering
\renewcommand{\arraystretch}{1.}
\footnotesize
\resizebox{0.9\textwidth}{!}{
\begin{tabular}{lccccc}
\toprule
  &  &  &  & \multicolumn{2}{c}{Color jitter ($p=0.2$)} \\
\cmidrule{5-6}
Category  & Vertical flip & Horizontal flip & Random rotation ($\pm5^\circ$) & Normal data & Generated data\\
\midrule
Bottle  & \checkmark & \checkmark & \checkmark & $0.05$ & $0.5$\\
Cable  & \ding{55} & \ding{55} & \checkmark & $0.05$ & $0.5$\\
Capsule  & \ding{55} & \ding{55} & \checkmark &$0.05$ & $0.5$\\
Carpet  & \checkmark & \checkmark & \checkmark &$0.05$ & $0.5$\\
Grid  & \checkmark & \checkmark & \checkmark & $0.05$ & $0.5$\\
Hazelnut  & \checkmark & \checkmark & \checkmark & $0.05$ & $0.5$\\
Leather  & \checkmark & \checkmark & \checkmark & $0.05$ & $0.5$\\
Metal Nut  & \ding{55} & \ding{55} & \checkmark & $0.05$ & $0.5$\\
Pill  & \ding{55} & \ding{55} & \checkmark & $0.05$ & $0.5$\\
Screw  & \checkmark & \checkmark & \checkmark & $0.05$ & $0.5$\\
Tile  & \checkmark & \checkmark & \checkmark & $0.05$ & $0.5$\\
Toothbrush  & \ding{55} & \checkmark & \checkmark & $0.05$ & $0.5$\\
Transistor  & \ding{55} & \checkmark & \checkmark & $0.05$ & $0.5$\\
Wood  & \checkmark & \checkmark & \checkmark & $0.05$ & $0.5$\\
Zipper  & \checkmark & \checkmark & \checkmark & $0.05$ & $0.5$\\
\bottomrule
\end{tabular}}
\caption{Overview of the dataset augmentation techniques applied during training to each of the sub-dataset present in MVTec AD, similar to the setting as relative works~\cite{MVloco}.}
\label{tab:augmentation_mvtec}
\end{table}

\begin{table}[!htbp]
\centering
\footnotesize
\resizebox{1\textwidth}{!}
{
\begin{tabular}{lccccc}
\toprule
  &  &  &  & \multicolumn{2}{c}{Color jitter ($p=0.2$)} \\
\cmidrule{5-6}
Category  & Vertical flip & Horizontal flip & Random rotation ($\pm5^\circ$) & Normal data & Generated data\\
\midrule
Breakfast Box  & \ding{55} & \ding{55} & \checkmark & $0.05$ & $0.5$\\
Screw Bag & \checkmark & \checkmark & \checkmark & $0.05$  & $0.5$\\
Pushpins & \checkmark & \checkmark & \checkmark & $0.05$  & $0.5$\\
Splicing Connectors & \checkmark & \checkmark & \checkmark & $0.05$  & $0.5$\\
Juice Bottle & \ding{55} & \ding{55} & \checkmark & $0.05$  & $0.5$\\
\bottomrule
\end{tabular}}
\caption{Overview of the dataset augmentation techniques applied during training to each of the sub-dataset present in MVTec LOCO, similar to the setting as relative works~\cite{MVloco}.}
\label{tab:augmentation_loco}
\end{table}

\begin{table}[!htpb]
\centering
\footnotesize
\begin{tabular}{lcccc}
\toprule
Dataset     &Detector level & \makecell{Practical size of\\receptive field} & input size & PatchDiff level\\
\midrule
MVTec AD                       & 34    & $34\times34$  & $256\times256$   & 5, 9, 13   \\
\midrule
\multirow{3}{*}{MVTec LoCo AD} & 34    & $34\times34$  & $256\times256$   & 5, 9, 13   \\
                               & 68    & $34\times34$  & $128\times128$   & 5, 9, 13   \\
                               & 136   & $34\times34$  & $64\times64$     & 9, 13, 17 \\
\bottomrule    
\end{tabular}
\caption{The level configures list for patch-level detectors.}
\label{tab: GRad_level_configs}
\end{table}

\subsubsection{Training Detail} We then present the configuration details of the patch-level detectors across various levels, as outlined in Table \ref{tab: GRad_level_configs}. In the case of the MVTec AD dataset, we exclusively train a level-$34$ patch-level detector for each sub-dataset. In addition, for MVTec LOCO, we developed three detectors — each corresponding to level-$34$, $68$, and $136$ within their respective sub-datasets. Concerning these levels, the images are resized to dimensions of $256\times256$, $128\times128$, and $64\times64$ respectively. This resizing strategy allows us to maintain the practical receptive field size of each detector at $34\times34$, while the effective receptive field relatively expands to $34\times34$, $68\times68$, and $136\times136$ for the level-$34$, $68$, and $136$ detectors respectively. Moreover, the level-$34$ and $68$ patch-level detectors employ contrastive images generated by level-$5$, $9$, and $13$ PatchDiffs, whereas the level-$136$ patch-level detector employs contrastive images generated by level-$9$, $13$, and $17$ PatchDiffs. This meticulous selection of contrastive images from varying levels PatchDiff further contributes to the detectors' adeptness in capturing diverse local anomaly patterns. Moreover, for our reweighting mechanism, we introduce a memory bank size of 512 for storing the normal features during the training phase. We train the detector for 2000 epochs using AdamW with a one-cycle learning rate scheduler and an initial learning rate of 1e-3.

\begin{table}[!htpb]
\centering
\footnotesize
\begin{tabular}{cccccc}
\toprule
Layer Name & Stride & Kernel Size & Number of Kernels & Padding & Activation \\
\midrule
Conv-1 & 2$\times$2 & 4$\times$4 & 64 & 0 & ReLU \\
Conv-2 & 2$\times$2 & 4$\times$4 & 128 & 0 & ReLU \\
Conv-3 & 1$\times$1 & 3$\times$3 & 256 & 0 & ReLU \\
Conv-4 & 1$\times$1 & 3$\times$3 & 512 & 0 & ReLU \\
Conv-5 & 1$\times$1 & 3$\times$3 & 256 & 0 & ReLU \\
Conv-6 & 1$\times$1 & 1$\times$1 & 256 & 0 & ReLU \\
Conv-7 & 1$\times$1 & 1$\times$1 & 256 & 0 & ReLU \\
Conv-8 & 1$\times$1 & 1$\times$1 & 1 & 0 & - \\
\bottomrule
\end{tabular}
\caption{Network architecture of our patch-level detector.}
\label{tab: arch_detector}
\end{table}

\begin{table}[!htpb]
\centering
\footnotesize
\begin{tabular}{cccccc}
\toprule
Layer Name & Stride & Kernel Size & Number of Kernels & Padding & Activation \\
\midrule
Conv-1 & 1$\times$1 & 1$\times$1 & 256 & 0 & ReLU \\
Conv-2 & 1$\times$1 & 1$\times$1 & 256 & 0 & ReLU \\
Conv-3 & 1$\times$1 & 1$\times$1 & 256 & 0 & ReLU \\
Conv-4 & 1$\times$1 & 1$\times$1 & 5780 & 0 & - \\
\bottomrule
\end{tabular}
\caption{Network architecture of MLP-based decoder for our regularization on features.}
\label{tab: arch_decoder}
\end{table}

\subsubsection{Network Architecture} Additionally, we illustrate the network architecture of our patch-level detector in Table~\ref{tab: arch_detector}, which outputs $1\times1$ anomaly score for an input patch size of $5\times34\times34$ pixels. Consequently, each individual patch-level detector encompasses around 2.9 million parameters, equipped with only 8 fully convolutional layers. Even though we integrate the comprehensive performance of three detectors for MVTec LOCO, the whole number of parameters is still only 8.7 million parameters, highlighting its lightweight structure when compared to prevailing backbone architectures as shown in first figure of our paper. Moreover, in Table~\ref{tab: arch_decoder}, we further illustrate the network architecture of MLP-based network for our regularization on features. We resize the output size $5780$ into $5\times34\times34$ pixels to achieve the reconstruction for the features encoded by our detectors. 

\subsection{Additional Experiment Results}
\subsection{Anomaly Generation}
Furthermore, as shown in Fig.~\ref{fig: mvtec_generation}, we present the samples of contrastive images generated by level-13 PatchDiff for MVTec AD.

\begin{figure*}[!h]
    \centering
\includegraphics[width=0.7\linewidth]{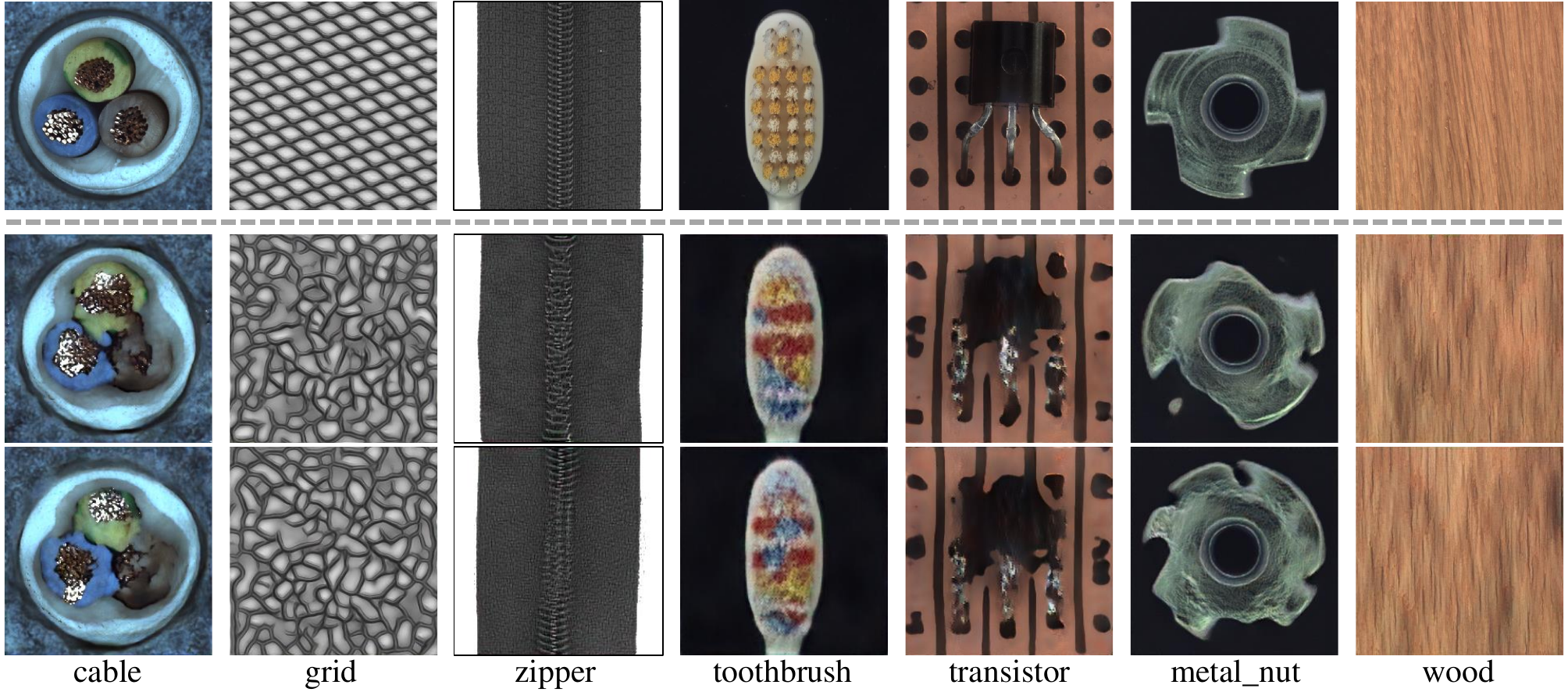}
    \caption{Contrastive images generated by level-13 PatchDiff for MVTec AD~\cite{MVTecAD}. } 
    \label{fig: mvtec_generation}
\end{figure*}

\begin{table*}[!h]
    \centering
    \renewcommand{\arraystretch}{1.2}
    \resizebox{\textwidth}{!}
    {
\begin{tabular}{cl|c|c|c|c|c|c|c|c}
\toprule
\multicolumn{2}{c|}{Category}     & IGD & PSVDD & FCDD & CutPaste &NSA & DRAEM & DSR & GRAD \\ \midrule
\multirow{5}{*}{Texture} 
& carpet & (94.7, 82.8 ) & (92.9, 92.6) & (96.0, - ) & (93.1, 98.3) & (95.5, 95.6) & (95.5,97.0) & (-, \textbf{100.}) & (\textbf{96.5}, 98.2) \\
& grid & (97.7, 97.8 ) & (94.6, 100.) & (91.0, - ) & (\textbf{99.9}, 97.5) & (99.2, 99.9) & (99.7, 99.9) & (-, \textbf{100.}) & (97.2, \textbf{100.}) \\
& leather & (99.5, 95.8) & (90.9, 98.6) & (98.0, - ) & (\textbf{100.}, 99.5) & (99.5, 99.9) & (98.6, \textbf{100.}) & (-, \textbf{100.}) & (98.8, \textbf{100.}) \\
& tile & (78.0, 99.1) & (97.8, 91.4) & (91.0, - ) & (93.4, 90.5) & (\textbf{99.3}, \textbf{100.}) & (99.2, 99.6) & (-, \textbf{100.}) & (95.4, \textbf{100.}) \\
& wood & (89.1, 94.6) & (96.5, 90.8) & (88.0, - ) & (\textbf{98.6}, 95.5) & (90.7, 97.5) & (96.4, \textbf{99.1}) & (-, 96.3) & (87.2, 98.3) \\
\midrule
\multirow{10}{*}{Object} 
& bottle & (92.2, \textbf{100.}) & (98.6, 98.1) & (97.0, - ) & (98.3, 97.6) & (98.3, 97.7) & (\textbf{99.1}, 99.2) & (-, \textbf{100.}) & (96.5, \textbf{100.}) \\
& cable & (84.7, 90.6) & (90.3, 96.8) & (90.0, - ) & (80.6, 90.0) & (96.0, 94.5) & (94.7, 91.8) & (-, 93.8) & (\textbf{98.4}, \textbf{99.3}) \\
& capsule & (\textbf{97.7}, 91.5) & (76.7, 95.8) & (93.0, - ) & (96.2, 97.4) & (97.6, 95.2) & (94.3, \textbf{98.5}) & (-, 98.1) & (97.1, 96.4) \\
& hazelnut & (98.0, 99.7) & (92.0, 97.5) & (95.0, - ) & (97.3, 97.3) & (97.6, 94.7) & (\textbf{99.7}, \textbf{100.}) & (-, 95.6) & (96.6, 98.1) \\
& metal nut & (92.6, 91.3) & (94.0, 98.0) & (94.0, - ) & (99.3, 93.1) & (98.4, 98.7) & (\textbf{99.5}, 98.7) & (-, 98.5) & (93.7, \textbf{100.}) \\
& pill & (97.3, 87.3) & (86.1, 95.1) & (81.0, - ) & (92.4, 95.7) & (\textbf{98.5}, \textbf{99.2}) & (97.6, 98.9) & (-, 97.5) & (98.1, 95.7) \\
& screw & (97.0, 82.5) & (81.3, 95.7) & (86.0, - ) & (86.3, \textbf{96.7}) & (96.5, 90.2) & (97.6, 93.9) & (-, 96.2) & (\textbf{99.2}, 96.0) \\
& toothbrush & (97.7, 99.7) & (\textbf{100.}, 98.1) & (94.0, - ) & (98.3, 98.1) & (94.9, \textbf{100.}) & (98.1, \textbf{100.}) & (-, 99.7) & (98.0, 99.7) \\
& transistor & (84.4, 90.6) & (91.5, 97.0) & (88.0, - ) & (95.5, 93.0) & (88.0, 95.1) & (90.9, 93.1) & (-, 97.8) & (\textbf{97.8}, \textbf{100.}) \\
& zipper & (96.7, 97.0) & (97.9, 95.1) & (92.0, - ) & (\textbf{99.4}, 99.3) & (94.2, 99.8) & (98.9, \textbf{100.}) & (-, \textbf{100.}) & (98.3, 99.7) \\
\midrule
\multicolumn{2}{c|}{Average} & (93.1, 93.4) & (92.5, 93.2 ) & (92.1, 95.7) & (95.2, 96.0) & (96.3, 97.2) & (\textbf{97.3}, 98.0) & (-, 98.2) & (96.8, \textbf{98.7}) \\
\bottomrule
\end{tabular}}
\caption{Anomaly detection performance on MVTec AD dataset~\cite{MVTecAD}. Both pixel-level (left) and image-level (right) AUROC results are shown in each column. The best results are in bold.}
\label{tab: mvtec_main_detail}
\end{table*}

\subsection{Anomaly Detection and Localization}
In the main body, we exclusively present the averaged performance comparison on MVTec AD. In this section, we extend our analysis to provide a detailed result of the anomaly detection and localization performance across each individual sub-dataset within MVTec AD, and display anomaly maps on MVTec AD in Fig.~\ref{fig: main_mvtec_ad_results}. As shown in Table~\ref{tab: mvtec_main_detail}, we compare GRAD to IGD~\cite{IGD}, PSVDD~\cite{PSVDD}, FCDD~\cite{FCDD}, CutPaste~\cite{CutPaste}, NSA~\cite{NSA}, DRAEM~\cite{DRAEM}, and DSR~\cite{DSR}, all of which are independent of pretrained feature extractors. It is easy to find GRAD achieves a strong detection and localization of anomalies.  

\begin{figure*}[!h]
    \centering
\includegraphics[width=0.9\linewidth]{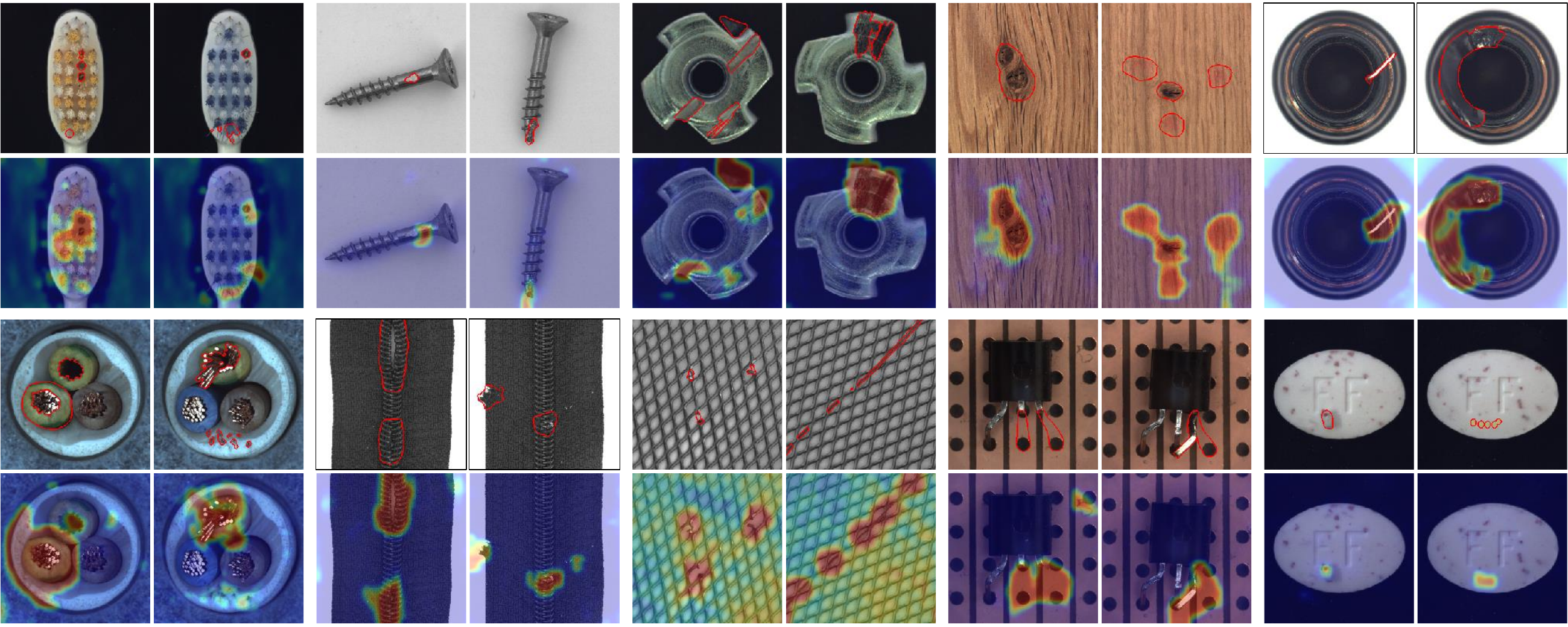}
    \caption{Defect localization results of GRAD on MVTec AD~\cite{MVTecAD}. } 
    \label{fig: main_mvtec_ad_results}
\end{figure*}





}

\end{document}